%% file: main.tex
\documentclass[runningheads]{llncs}

\usepackage{eccv}

\usepackage{algorithm}
\usepackage{algorithmic}
\usepackage{xcolor,colortbl}
\definecolor{Gray}{gray}{0.85}
\definecolor{LightCyan}{rgb}{0.88,1,1}
\newcolumntype{a}{>{\columncolor{Gray}}c}
\usepackage{amsmath}
\usepackage{xcolor}
\usepackage{amsfonts}
\usepackage{pgfplots}
\usepackage{subcaption}
\usepackage{booktabs}
\usepackage{bm}
\usepackage{caption} 
\usepackage{multirow}
\usepackage{multicol}
\usepackage{xcolor,colortbl}

\usepackage[misc]{ifsym}

\usepackage{eccvabbrv}

\usepackage{graphicx}
\usepackage{booktabs}

\makeatletter
\def\blfootnote{\gdef\@thefnmark{}\@footnotetext}
\makeatother

\usepackage[accsupp]{axessibility}  

\usepackage{hyperref}

\usepackage{orcidlink}

\begin{document}

\title{GKGNet: Group K-Nearest Neighbor based Graph Convolutional Network for Multi-Label Image Recognition} 

\titlerunning{GKGNet}

\author{
    Ruijie Yao\inst{1,4} \orcidlink{0009-0007-4736-9453} \and
    Sheng Jin\inst{2,4}\orcidlink{0000-0001-5736-7434} \and
    Lumin Xu\inst{3}\orcidlink{0000-0003-2125-2760} \and
    Wang Zeng\inst{4}\orcidlink{0000-0003-1562-6332} \and
    Wentao Liu\inst{4}\orcidlink{0000-0001-6587-9878} \and
    Chen Qian\inst{1,4}\orcidlink{0000-0002-8761-5563} \textsuperscript{\Letter} \and
    Ping Luo\inst{2,5}\orcidlink{0000-0002-6685-7950} \and
    Ji Wu\inst{1} \orcidlink{0000-0001-6170-726X} \textsuperscript{\Letter}
}

\authorrunning{R. Yao et al.}

\institute{$^{1}$ Tsinghua University \quad 
$^{2}$ The University of Hong Kong \quad
$^{3}$ The Chinese University of Hong Kong \quad
$^{4}$ SenseTime Research and Tetras.AI \quad
$^{5}$ Shanghai AI Laboratory \\
\email{\{yrj21@mails, qianc18@mails, wuji\_ee@mail\}.tsinghua.edu.cn}}
\maketitle

\begin{abstract}
Multi-Label Image Recognition (MLIR) is a challenging task that aims to predict multiple object labels in a single image while modeling the complex relationships between labels and image regions. Although convolutional neural networks and vision transformers have succeeded in processing images as regular grids of pixels or patches, these representations are sub-optimal for capturing irregular and discontinuous regions of interest. In this work, we present the first fully graph convolutional model, Group K-nearest neighbor based Graph convolutional Network (GKGNet), which models the connections between semantic label embeddings and image patches in a flexible and unified graph structure. To address the scale variance of different objects and to capture information from multiple perspectives, we propose the Group KGCN module for dynamic graph construction and message passing. Our experiments demonstrate that GKGNet achieves state-of-the-art performance with significantly lower computational costs on the challenging multi-label datasets, \ie MS-COCO and VOC2007 datasets. Codes are available at \url{https://github.com/jin-s13/GKGNet}.

\blfootnote{\Letter~: Corresponding authors.}
\end{abstract}

\section{Introduction}
\label{sec:introduction}

Multi-label image recognition (MLIR) (also referred to as multi-label classification) is a fundamental task in computer vision, which aims to predict a set of labels of a single image. Compared with single-label image recognition, MLIR is more challenging due to its combinatorial nature. It has received increasing attention because of its broad real-world applications, such as human attribute recognition~\cite{li2016human} and scene understanding~\cite{scene-understanding}. 

\begin{figure}[ht]
\centering
\includegraphics[width=0.7\textwidth]{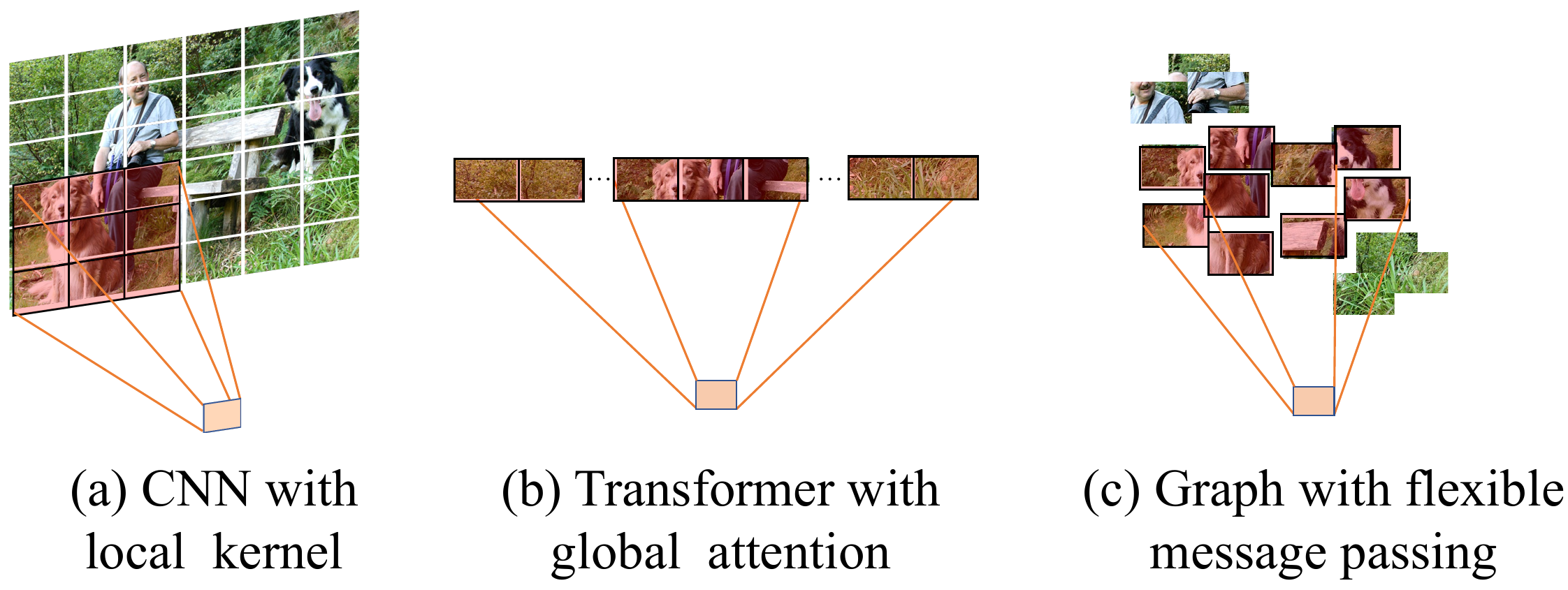}
\caption{Illustration of feature extraction in CNN, vision transformer, and graph convolutional network (GCN).
(a) CNN excels at processing continuous regions but struggles with irregular regions of interest.
(b) Vision transformers handle complex regions of interest but introduce redundant interference from the background.
(c) GCN constructs connections between the destination node and multiple objects of interest distributed in different spatial locations.
}
\label{fig:introduce}
\end{figure}

Modeling the relationships between each target label and the corresponding image regions is critical yet challenging for MLIR. The regions that relate to a certain label may be complicated and even discontinuous. For example, to recognize the presence of dogs in an image, multiple regions need to follow with interest, since there may exist multiple dogs. Convolutional neural networks (CNN)~\cite{resnet101,srn,pedestrian-jia2021spatial} regard the image as grids of pixels and apply sliding convolutional kernels according to spatial positions. As shown in Fig.~\ref{fig:introduce}  (a), CNN process continuous regions well but  struggles to fit irregular regions of interest. In contrast, vision transformers~\cite{mltr} treat the image as a sequence of patches and extract the visual features from the patches via attention. As shown in Fig.~\ref{fig:introduce}  (b), vision transformers can handle complicated regions of interest by capturing the whole image features. However, interactions with the background increase costs and affect performance. For small-size target objects, most patches belong to the background thus the sum of their attention scores cannot be ignored.
Graph methods like Vision GNN~\cite{vig} views visual patches as nodes, forming a flexible approach to extract features by representing images as graphs.
As shown in Fig.~\ref{fig:introduce}  (c), the graph representation constructs the connections according to the semantic meanings and focuses on the regions of interest distributed in different spatial locations.

In this work, we propose the first fully graph convolutional network (GCN) for the task of MLIR, termed \textbf{G}roup \textbf{K}-nearest neighbor based \textbf{G}raph convolutional \textbf{Net}work (GKGNet). GKGNet regards both the image patches and target labels as graph nodes and processes them in a unified graph structure.
GKGNet constructs two distinct graphs: a cross-level graph that models the label-object relationship between target labels and image patches, and a patch-level graph that processes and updates the image features between patches. Thus, GKGNet enables the adaptive integration of features from patches of interest, even in the presence of irregular and discontinuous regions, leading to the effective updating of the unified graph representations of visual features and label embeddings.

Graph construction is the key to the success of GCN.
The K-nearest neighbor (KNN) graph, where there is an edge between each node and its $K$ most similar nodes, is one of the most widely used graph structures~\cite{vig}. However, this approach is sub-optimal for the following reasons.
First, the number of neighbors $K$ controls the size of the area where region features will be extracted and aggregated. Large $K$ will lead to feature over-smoothing and involve interference of the invalid background, while small $K$ will affect feature extraction and message passing. A fixed $K$ cannot adaptively handle objects with different scales. 
Second, this method finds the neighbors by measuring the distance between the destination node and the candidate source nodes. It is difficult to define a unique measurement of distance to represent the rich dimensions of ``high-level'' labels.
 
Therefore, we propose the Group KNN-based GCN (Group KGCN) module, which splits node features into multiple groups and constructs connections among each group.
By designing each group to potentially connect to different source nodes, the Group KNN approach allows a destination node to interact with a dynamic number of source nodes and handle objects with different scales.
Also, this dynamic connectivity enables the model to capture information from multiple perspectives, which is crucial for MLIR. For instance, recognizing the class ``zebra" requires consideration of both shape and appearance features. Furthermore, MLIR involves capturing multi-label correlations, where the context surrounding the target objects is also important for accurate recognition~\cite{ssgrl,ml-gcn,kssnet,add-gcn,ms-cma,tdrg}. Therefore, by allowing for flexible and varied connectivity patterns, the Group KNN approach is able to model these complex relationships between image regions and labels.

Experiments on two well-known benchmarks, \ie MS-COCO and VOC2007, verify the effectiveness of the proposed method. Our method achieves new state-of-the-art performance with significantly lower computational costs. 

This paper's contribution can be summarized as follows:
    \begin{itemize}
        \item We introduce GKGNet, the first fully graph convolutional model for multi-label image recognition, which builds unified graph representations for visual features and label embeddings, explicitly modeling the flexible relationship between labels and regions of interest.
        
        \item To handle scale variance and capture diverse perspectives, we propose the Group KGCN module for dynamic graph construction and message passing.
        
        \item Our proposed model is validated on popular benchmark datasets, including MS-COCO and VOC2007. It consistently outperforms previous state-of-the-art approaches with lower computational complexity.
    \end{itemize}

\begin{figure*}[ht]
\begin{center}
\includegraphics[width=0.98\linewidth]{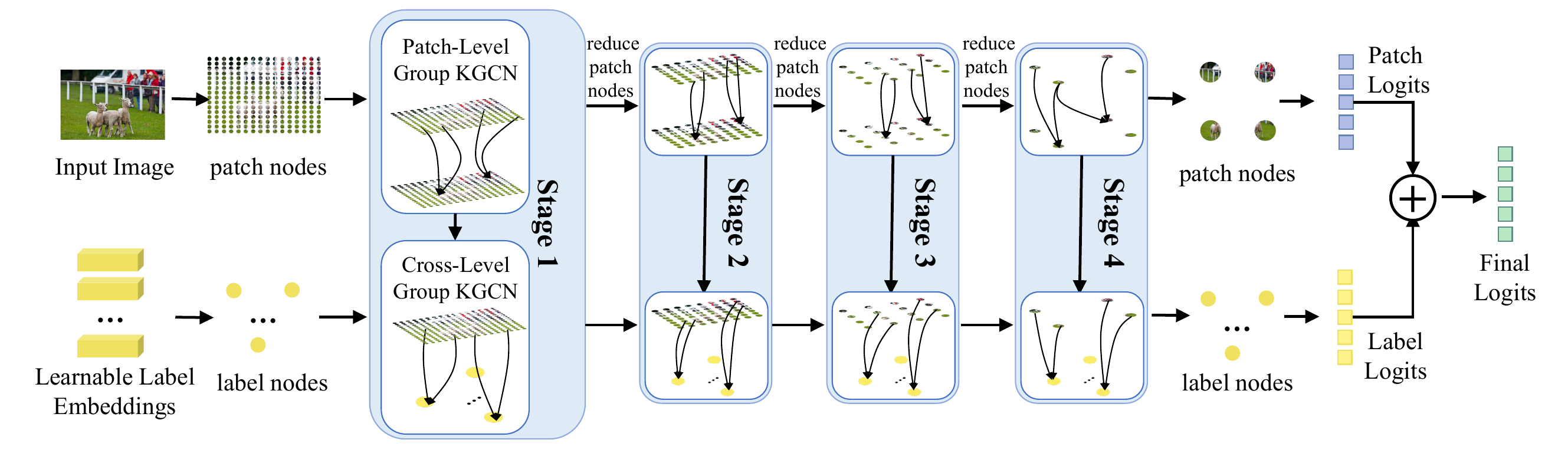}
\end{center}
\caption{\textbf{Overview of GKGNet.} 
GKGNet splits the input image into a set of patch nodes, and regards the learnable label embeddings as label nodes. Four-stage network is applied to process the patch nodes and label nodes in the unified graph structure.
The number of patch nodes is reduced after each stage to extract multi-scale visual features. At each stage, the patch nodes are first updated via Patch-Level Group KGCN modules, and then Cross-Level Group KGCN modules updates the label nodes by building the connections between target labels and image regions of interest. The output patch nodes and label nodes of the last stage are combined for multi-label prediction. 
}
\label{fig:framework}
\end{figure*}

\section{Related Work}

\subsection{Multi-Label Image Recognition}

Multi-label image recognition (MLIR) (also referred to as multi-label classification) is an extension of image classification, which aims to predict multiple class labels for an image. 
Several approaches~\cite{resnet101,srn,pedestrian-jia2021spatial} design powerful Convolutional Neural Network (CNN) architectures for the task of MLIR. CNN extracts the visual features of the input image and predicts multiples labels according to the same features. Instead of explicitly capturing the regions of interest, these methods treat each label equally and simply consider MLIR as multiple binary classification tasks.
Recently, some works~\cite{c-tran,liu2021query2label,mltr} explore to use transformers for the task of MLIR to capture semantic relationships among the labels.
C-Tran~\cite{c-tran} uses a transformer encoder to process image features and label embeddings simultaneously, while Q2L~\cite{liu2021query2label} employs a transformer decoder to explicitly exchange information between target labels and image features.
These methods predict labels using visual features of the whole image, leading to redundant background interference and performance hindrance. Moreover, such designs demand high computational costs, particularly as image resolution increases.

There are several works~\cite{ml-gcn,kssnet,add-gcn,tdrg,chen2020knowledge,chen2021learning} applying graph convolutional network (GCN) for MLIR.
They use CNN for image representation learning and only use GCN to model the relationship between multiple labels. However, these methods do not explore the important connections between the global classification label and spatial visual regions. 
Moreover, the image representations learned by CNN are not well-aligned with the semantic embeddings of labels processed by GCN, which hinders message passing.
To deal with the above problems, we instead propose a fully graph convolutional network, termed GKGNet, to capture the relationship between target labels and the spatial areas in a unified graph representation. GKGNet adaptly searches the regions of interest for each target label in the flexible graph structure. It is able to fit in irregular areas and capture complex correlations without introducing any computational cost.

\subsection{Graph Convolutional Network}
Graph is a versatile and flexible data structure and it can process any kind of data that can be converted into a set of nodes and edges. They can represent both non-Euclidean data, such as social networks, and Euclidean data, such as images~\cite{gcn-review}. Graph convolutional networks (GCN)~\cite{henaff2015deep} are introduced to learn graph representations using convolutional operations on the graph's Laplacian matrix based on spectral graph theory. GCN is highly effective for message passing and correlation modeling in computer vision tasks, such as multi-label classification~\cite{ml-gcn}, relationship proposal generation~\cite{wang2020affinity}, scene graph generation~\cite{Scene-graph-generation}, and human pose estimation~\cite{jin2020differentiable}. Recently, a GCN-based backbone network called ViG~\cite{vig} has been proposed to directly convert images into graph structures to learn visual representations. This approach divides an image into patches and considers them as nodes in a graph. Edges are constructed by identifying K-nearest neighbors (KNN) for each node, and effective message passing captures the spatial relationships between visual patches. However, the KNN-based approach limits the destination node to aggregate information from a fixed number of source nodes, which may not be appropriate for tasks involving target labels with regions of different sizes and shapes. To address this limitation, we propose the Group KGCN module, which can adapt to different numbers of neighbors based on practical requirements.

\section{Method}

\subsection{Overview}
\label{sec:overview}
We propose a fully graph convolutional network (GCN), termed Group K-nearest neighbor based Graph convolutional Network (GKGNet), for the task of multi-label image recognition. The overview of GKGNet is illustrated in Fig.~\ref{fig:framework}. 
Inspired by ViG~\cite{vig},
the input image is divided into $N$ image patches, and each patch is transformed into a visual feature vector with dimension $C$ by a fully-connected layer. 
These feature vectors are represented as patch nodes in our graph representations.
The learnable label embeddings are viewed as label nodes, which have the same feature dimension $C$ as the patch nodes.
The patch nodes and label nodes are processed by four hierarchical stages, and the number of patch nodes is reduced after each stage to extract multi-scale visual features. 
Each stage consists of two kinds of Group KGCN modules, \ie Patch-Level Group KGCN module and Cross-Level Group KGCN module. Patch-Level Group KGCN module serves as image feature extractor to capture the spatial relationship between visual features, and Cross-Level Group KGCN module establishes the cross-correlation between target labels and the related regions of interest.
In order to construct flexible connections between nodes, Group KGCN module allows each node to correlate with different number of patch nodes according to practical needs.
Finally, the classifier is applied to the output patch nodes and label nodes, and both of them are taken into consideration for class possibility prediction.

\subsection{Group KGCN Module}
\label{sec:group_knn}

Our GKGNet paradigm entails the representation of both patch and label features in a graph form, wherein edges are established among neighboring nodes to facilitate message passing. A conventional technique for graph construction involves identifying the K-nearest neighbors (KNN) of the destination node, as depicted in Fig.~\ref{fig:group_knn}  (a), and subsequently updating the destination node's features based on those of its neighbors, \ie source nodes. However, such a methodology restricts itself to a fixed number of source nodes and may not be optimal for the task of MLIR, because the target label may be associated with regions of varying sizes and different number of source nodes should be connected.
To address this issue, we propose Group K-nearest neighbors (Group KNN), which provides greater flexibility in constructing the graph structure, for Group KNN-based GCN (Group KGCN) module, enabling effective message passing.

\subsubsection{Group KNN.}
 Group KNN based graph construction divides both source nodes $\mathbb{S} = \{S_{1}, S_{2}, \cdot\cdot\cdot, S_{N_S}\} \in \mathbb{R}^{N_S\times C}$ and destination nodes $\mathbb{D} = \{D_{1}, D_{2}, \cdot\cdot\cdot, D_{N_D}\} \in \mathbb{R}^{N_D\times C}$ into $G$ groups of sub source nodes and sub destination nodes by splitting feature dimensions, where $C$ denotes the dimension of the destination node and source node, and $N_S$ and $N_D$ are numbers of source nodes and destination nodes respectively.
 Each sub destination node selects to connect with sub source nodes within the same group.
 For group $g$, the $i$-th sub destination node $D_{ig} \in \mathbb{R}^{\frac{C}{G}}$ ($i \in [1, N_D], g \in [1, G]$) searches for the K-nearest neighbors among the corresponding sub source nodes $\mathbb{S}_{g} = \{S_{1g}, S_{2g}, \cdot\cdot\cdot, S_{N_{S}g}\} \in \mathbb{R}^{N_S\times \frac{C}{G}}$. Cosine similarity is used for neighborhood search.

 \begin{figure} 
\centering
\includegraphics[width=0.5\textwidth]{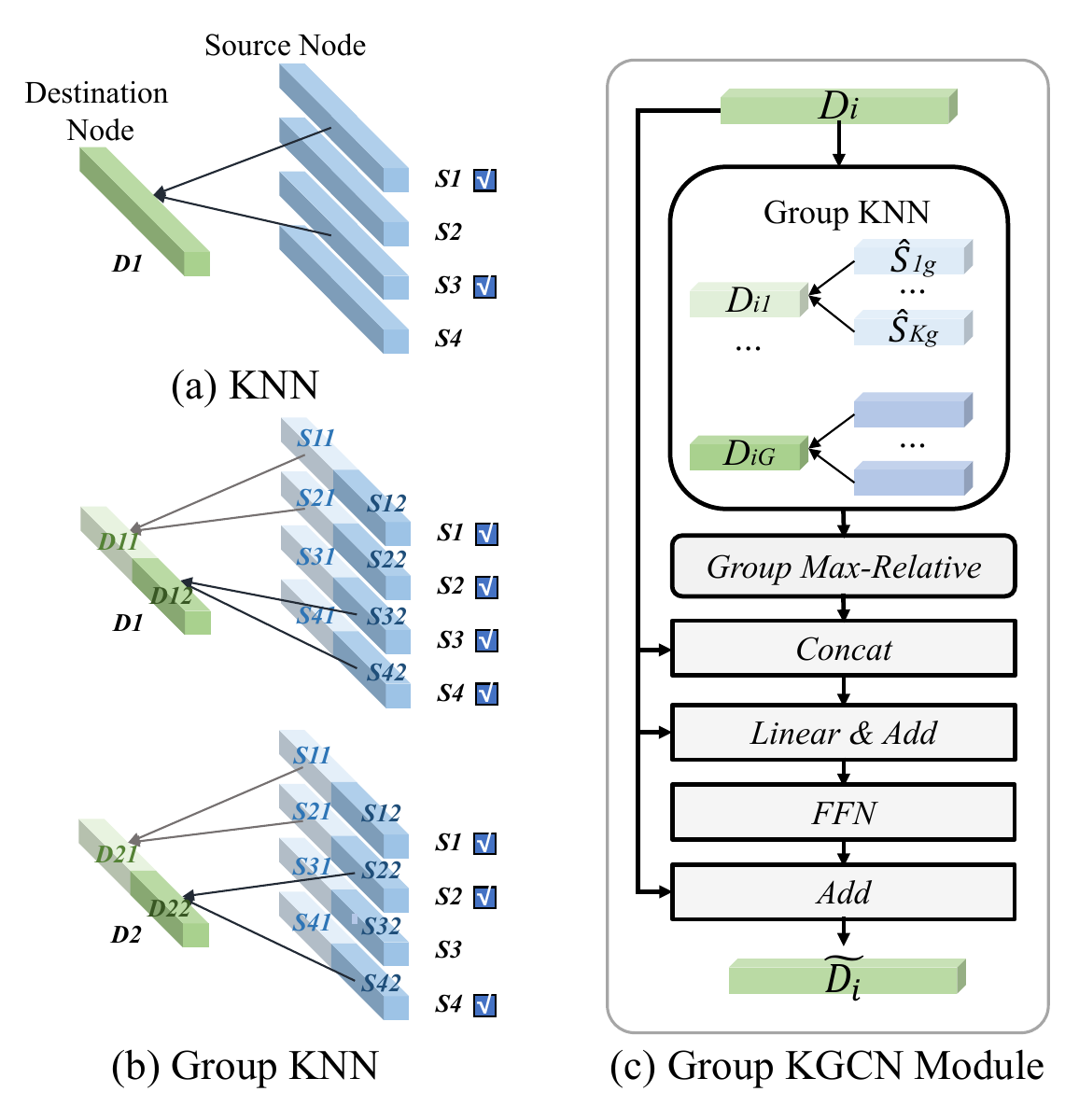}
\caption{\textbf{Illustration of Group KGCN.}  (a) Traditional KNN based graph construction (K=2).  (b) Group KNN based graph construction (G=2, K=2). The blue check marks indicate the source nodes are selected. (c) Structure of Group KGCN module.}
\label{fig:group_knn}
\end{figure}

The utilization of distinct groups allows the destination node to establish connections with a variable number of source nodes, ranging from $K$ to $K\times G$. As shown in Figure~\ref{fig:group_knn} (b), the sub destination node $D_{11}$ links to the sub source nodes $S_{11}$ and $S_{21}$, while the sub destination node $D_{12}$ connects with the sub source nodes $S_{31}$ and $S_{41}$. This enables the destination node $D_1$ to interact with four source nodes: $S_1$, $S_2$, $S_3$, and $S_4$. This scenario typically occurs when the destination node concerns a broader area of interest, such as a large target object.
Conversely, when the neighbors of distinct groups overlap, the number of selected source nodes decreases. For instance, both sub destination nodes of $D_2$ connect with the source nodes $S_2$, resulting in an interaction with only three source nodes: $S_1$, $S_2$, and $S_4$. This scenario typically occurs when the destination node involves only small regions of interest, such as a small target object, thereby helping to circumvent irrelevant information during message passing.

\subsubsection{Module structure.}
As shown in Fig.~\ref{fig:group_knn} (c), the Group KGCN module conducts message passing between the destination node and source nodes after the graph is constructed via Group KNN.
The sub destination node $D_{ig}$ is updated according to its K-nearest neighbor source nodes $\{\hat{\mathcal{{S}}_{1g}}, \hat{\mathcal{{S}}_{2g}}, \cdot\cdot\cdot, \hat{\mathcal{{S}}_{Kg}}\} \subseteq \mathbb{S}_{g}$ via group max-relative graph convolution. $\hat{S}$ represents the selected nearest neighbors.

Group max-relative graph convolution is an extension of~\cite{li2019deepgcns}. For each sub destination node, it aggregates features of its group-wise neighbors by maximizing every dimension of the relative feature $D_{ig}-\hat{\mathcal{{S}}_{kg}}$ as follows:

\begin{equation}
\begin{aligned}
    D^{'}_{ig}=  \ max(\{{D_{ig}}- \hat{\mathcal{{S}}_{kg}} | k \in [1, K]\}).
\end{aligned}
\end{equation}

Then all the updated sub destination nodes are processed to output the renewed destination node $\widetilde{D_i}$ as follows:

\begin{equation}
\widetilde{D_i}=  D_{i}+FFN(D_{i} + Linear(   Concat({D_i}, \{D^{'}_{ig}\}))),
\label{global formula}
\end{equation}
\noindent where $Concat$ indicates the concatenation operation, and $Linear$ is a  fully connected layer. $FFN$ denotes the feed-forward network. The original destination node $D_i$ and its $G$ sub-destination nodes processed by group max-relative graph convolution are concatenated together for destination node feature update. The residual structure is used to relieve the over-smoothing problem.

\subsubsection{Computation cost analysis.}
Our Group KGCN module improves model accuracy without adding computational costs. Traditional KNN-based GCNs have a distance calculation cost of \(O(N_S \times N_D \times C)\) for \(N_S\) source nodes, \(N_D\) destination nodes, and dimension \(C\). The Group KGCN module has a cost of \(O(G \times N_S \times N_D \times \frac{C}{G}) = O(N_S \times N_D \times C)\), which is the same.


\subsection{Patch-Level and Cross-Level Group KGCN}
\label{sec:GKG-block}
Each GKGNet stage consists of two types of Group KGCN modules. Patch-Level Group KGCN module updates visual features within patches and Cross-Level Group KGCN module builds connections between labels and patches.

We apply Patch-Level Group KGCN modules, where both destination and source nodes are patch nodes, to extract image features. 
This graph-based approach allows modeling semantic relationships among  regions. Patch-Level Group KGCN modules capture the parts of an object or multiple objects of the same category distributed in different locations for more robust visual representations.

Besides the relationship among patch-level visual features, we take into consideration the cross-level label-patch correlation. We construct Cross-Level Group KGCN modules that enable message passing from patch nodes (source nodes) to label nodes (destination nodes),  to learn label embeddings from the region features. The graph representation provides a flexible way to extract global context information, which are important for the task of multi-label image recognition (MLIR) where the target objects may be irregular and discontinuous. Group KNN based graph construction enables each label to capture objects of different scales and multi-label correlations. By constructing the unified graph representation for labels and patches, the cross-level graph effectively benefit label recognition from the regions of interest in the same representation space.

\subsection{Classifier and Loss Function}
The classifier utilizes both patch nodes and label nodes outputted by the last stage for MLIR. The final scores are:
\begin{equation}
    {{Y}}=Sigmoid({{Y}}_{x_p}+{{Y}}_{x_l}),
\end{equation}
 where ${{Y}}_{x_p} \in \mathbb{R}^{L}$ and ${{Y}}_{x_l} \in \mathbb{R}^{L}$ are the logits predicted by patch nodes and label nodes using the fully connected layers respectively.
 $L$ is the number of target categories.
Label smooth loss $\mathcal{L}_{smooth}$~\cite{labelsmooth} and asymmetric loss $\mathcal{L}_{asy}$~\cite{asl} are applied to supervise the training process.
The total training loss is $\mathcal{L}=\mathcal{L}_{smooth}+\mathcal{L}_{asy}$.

\input{tables/table_coco.tex}
\section{Experiments}
We evaluate our model on several benchmark datasets, including MS-COCO~\cite{ms-coco} and Pascal VOC~\cite{voc}. We follow common practice~\cite{kssnet,c-tran,add-gcn} to report the average of Overall Recall (OR), Overall Precision (OP), Overall F1-score (OF1), per-Class Recall (CR), per-Class Precision (CP), per-Class F1-score (CF1) and the mean Average Precision (mAP) as the evaluation metrics. OF1, CF1 and mAP are emphasized as the most important metrics as OR, OP, CR and CP are easily affected by the classification threshold.
We set the threshold as 0.5 for recall, precision, and F1 score in our experiments.

\subsection{Implementation Details}

Numbers of Patch-Level Group KGCN modules are set as $[2, 2, 6, 2]$, and numbers of Cross-Level Group KGCN modules are $[1,1,1,1]$ for four hierarchical stages respectively. The feature dimensions $C$ for each stage are set as $[80,160,400,640]$.
We choose $K=9$ for the number of neighbors and $G=2$ for the number of groups in all the Group KGCN modules. 
ImageNet-1K~\cite{deng2009imagenet} pre-trained weights of Pyramid ViG-S~\cite{vig} are used for Patch-Level Group KGCN modules initialization, 
while Cross-Level Group KGCN modules and learnable label embeddings are randomly initialized.
During training, we utilize AdamW optimizer~\cite{adamw} with a weight decay rate of $0.05$ and momentum of 0.9 to train GKGNet for 80 epochs. The initial learning rate is $1\times10^{-4}$ with a linear warm-up of $1\times10^{-3}$.
We reduce the learning rate by a factor of 0.1 at the 5th and 50th epoch for Pascal VOC dataset, and at the 10th and 50th epoch for MS-COCO~\cite{ms-coco} dataset respectively.
All experiments are based on MMClassification~\cite{mmclassification}.

\subsection{Comparisons on MS-COCO Dataset} 

MS-COCO~\cite{ms-coco} is one of the most widely used benchmark datasets for multi-label classification. It contains 122,218 images that are composed of 82,081 training images and 40,137 validation images. There exist 80 kinds of object labels in total, with 2.9 labels on average for each image.
We report results on three commonly used input resolution settings on the dataset: $224 \times 224$, $448 \times 448$, and $576 \times 576$. All methods apply the models pre-trained on ImageNet-1K~\cite{deng2009imagenet} as the common practice. For our GKGNet, only Patch-Level Group KGCN uses pre-trained weights, while other parts use random initialization.
{For fair comparisons, we do not report results that apply much larger models (\eg Swin Large with 197M parameters) and extensive pre-training data (\eg ImageNet21K), which involves a different experiment setting from ours. }

As shown in Table~\ref{tab:ms-coco-table}, our method achieves the best performance on all the input resolution settings in terms of mAP, CF1, and OF1.
Compared with transformer based methods (\eg C-Tran~\cite{c-tran} and Q2L~\cite{liu2021query2label}), the flexibility of graph structure enables the model to focus only on the regions of interest and to avoid background interference, which saves computational cost and improves the performance. For example, GKGNet outperforms Q2L by $1.2\%$ mAP with half the FLOPs under the $576\times 576$ resolution setting. Kindly note that Q2L also utilizes Exponential Moving Average (EMA) technique that maintains moving averages of the trained model parameters to improve training robustness and model performance. 
ML-GCN~\cite{ml-gcn}, KSSNet~\cite{kssnet}, MS-CMA~\cite{ms-cma}, ADD-GCN~\cite{add-gcn}, SSGRL~\cite{ssgrl}, and TDRG~\cite{tdrg} are GCN based models. We find that our proposed GKGNet outperforms them by a large margin, which demonstrates the superiority of the unified design of patch-level and cross-level graph-based interaction, and the proposed Group KGCN modules that construct the graph structure adaptively according to the scale of target objects. Under the $448\times 448$ resolution setting, our model improves upon 
ML-GCN, KSSNet, MS-CMA, and TDRG by
$3.7\%$, $3.0\%$, $2.9\%$, and $2.1\%$\ mAP respectively.
We also compare with the model using the transformer backbone (IDA-SwinS(H)~\cite{liu2022causality}). Our method shows consistent superiority over the transformer backbone at various resolutions. For example, under the $448\times 448$ resolution setting, GKGNet outperforms IDA-SwinS(H) (86.7 mAP v.s. 85.5 mAP).

\begin{table}[htb]
\caption{We compare GKGNet with state-of-the-art methods using the same feature extractor ViG on MS-COCO ($448 \times 448$ input size) $^\dagger$ means using model EMA.
}
\scriptsize 
\centering
\setlength\tabcolsep{12pt}
\resizebox{0.8\linewidth}{!}{%
\begin{tabular}{l|a|c|c|c|c}
\hline
    \multirow{2}{*}{} &
  \multicolumn{3}{c|}{All} &
  \multicolumn{2}{c}{Top3} 
 \\ \cline{2-6}\cline{2-6}
 & mAP   & CF1    & OF1  & CF1 & OF1  \\ 
\hline
     {ViG~\cite{vig}}    &82.5   &76.6   &80.0     &72.7     &76.2\\
    + {C-Tran~\cite{c-tran}}   &84.5   &79.2   &80.8     &75.6   &77.0 \\
    + {Q2L~\cite{liu2021query2label}}   &85.5   &77.1   &78.2     &74.9     &76.8 \\
    + {Q2L~\cite{liu2021query2label}$^\dagger$}   &85.9&   79.2&   80.8&     76.1&     77.9\\ \hline
     {GKGNet}  & \textbf{86.7} & \textbf{81.5}  & \textbf{83.3} & \textbf{77.0} & \textbf{78.8}   \\
     \hline
\end{tabular}%
}
\label{tab:ms-coco-fair-table}
\end{table}

\subsubsection{Same backbone.} 
To decouple the effect of image feature extractor, we apply ViG to state-of-the-art methods (C-Tran and Q2L) using official codes and recommended training settings. The result of ViG is also reported as the baseline by replacing the final linear layer as ~\cite{resnet101}.
The same ImageNet-1K pre-trained weights are applied to all the models. The results of Q2L with and without the Exponential Moving Average (EMA) technique are both reported. As shown in Table~\ref{tab:ms-coco-fair-table}, our proposed GKGNet improves upon ViG baseline by 4.2\% mAP, and outperforms C-Tran and Q2L by a clear margin, which demonstrates the effectiveness of our specific design of GKGNet for the task of MLIR.

\subsection{Comparisons on Pascal VOC 2007 Dataset} 
Pascal VOC 2007~\cite{voc} is a commonly used benchmark dataset for MLIR, which contains 20 label categories. It has 9,963 images in total, and is divided into a train-val dataset (5,011 images) and a test dataset (4,952 images). The model is trained on the train-val dataset and evaluated on the test dataset as the common settings~\cite{asl,add-gcn,ssgrl,liu2021query2label}. For fairness, we follow the previous works~\cite{ssgrl,add-gcn} to pre-train the model on the MS-COCO and report results at $576\times 576$ resolution.

\input{tables/table_else.tex}

As shown in Table~\ref{tab:voc2007-table}, our approach achieves $0.7\%$ mAP improvement (96.8\% vs 96.1\% mAP) upon the previous state-of-the-art method Q2L. GKGNet consistently outperforms the previous methods on 14 of 20 categories, and achieves the competitive results on the remaining 6 categories. We also notice that the gain is 
more significant for these challenging categories, \eg chair and table.

\begin{table*}[tb]
\centering
\begin{minipage}[t]{0.45\textwidth}
\centering
 \caption{\textbf{Effect of model components in GKGNet.} 
    The experiments are conducted on MS-COCO ($448 \times 448$ input size).
    P, C, and G represent Patch-Level Graph, Cross-Level Graph, and Group KNN, respectively.}
    \label{tab:ablation}
    \centering
    \setlength{\tabcolsep}{7pt}
    \resizebox{0.95\linewidth}{!}{%
    \renewcommand{\arraystretch}{1.25}
    \begin{tabular}{c|c|c||c|c|c}
\hline
   {P} & {C} &{G} &   {mAP} &  {CF1} &  {OF1}\\
\hline
    &  &  & 79.9  & 74.6  & 78.7 \\
   \textbf{\checkmark} &  &  & 82.5  & 76.6  & 80.0 \\
   \textbf{\checkmark} & \textbf{\checkmark}  & & 85.5  & 80.4  & 82.6 \\ 
   \textbf{\checkmark}  & \textbf{\checkmark} & \textbf{\checkmark} & \textbf{86.7} & \textbf{81.5} & \textbf{83.3} \\
        \hline
\end{tabular}
}
\end{minipage}
\hspace{1.5mm}
\begin{minipage}[t]{0.45\textwidth}
\centering
\caption{\textbf{Effect of Group KNN on general classification}. Top-1 accuracy of the original Pyramid ViG-Tiny and the one enhanced with our Group KNN are reported on general classification datasets ($448 \times 448$ input size). }
\centering
\resizebox{0.96\linewidth}{!}{%
\begin{tabular}{l|c|c}
\hline
& ImageNet-1K  & CIFAR-10
\\
\hline
Pyramid ViG-Ti   & 78.2     & 94.6   \\
 \textbf{+} Group KNN  & \textbf{79.3} & \textbf{94.9}     \\
 \hline
& CIFAR-100  & Flowers   \\
\hline
Pyramid ViG-Ti    &74.4 &  83.6  \\
 \textbf{+} Group KNN   & \textbf{76.5} &\textbf{87.2}    \\
\hline
\end{tabular}%
}
\label{tab:imagenet}
\end{minipage}
\end{table*}

\subsection{Ablation Study}

\subsubsection{Effect of model components.} 
In GKGNet, the Patch-Level and Cross-Level Group KGCN modules model intricate object-object and label-object relationships,while Group KNN adaptively constructs graphs. Table~\ref{tab:ablation} validates these components individually. The first line shows the result of ResNet-50, which has comparable parameters to our model. Patch-Level graphs capture spatial relationships, achieving a 2.6\% gain. The Cross-Level graphs use unified representations of patch features and label embeddings to extract specific visual features for each target label, resulting in a 3.0\%  gain. Group KNN based graph construction handles scale variances, enhances feature diversity, and captures label correlations, leading to a significant improvement (86.7\% vs. 85.5\% mAP).

Additionally, Group KNN demonstrates potential for general image recognition. We apply Group KNN to the pyramid ViG-Tiny~\cite{vig}, and maintain fairness with identical training settings and hyper-parameters across all datasets. Training details are available in Supplementary.
As shown in Table~\ref{tab:imagenet}, our proposed Group KNN achieves a substantial improvement in top-1 accuracy on ImageNet-1K, CIFAR-10, CIFAR-100~\cite{krizhevsky2009learning}, and Flowers~\cite{nilsback2008automated} datasets, while maintaining the same model parameters and computational complexity (10.7M parameters and 1.7B FLOPs). The significant performance gain serves as strong evidence supporting the superiority of Group KNN based graph construction.

\begin{table*}[tb]
\centering
\begin{minipage}[t]{0.47\textwidth}
\centering
\caption{\textbf{Effect of object scales}.  We report  mAP for varying object sizes on MS-COCO with $448 \times 448$ input size.
}
\scriptsize 
\centering
\resizebox{0.99\linewidth}{!}{%
\setlength\tabcolsep{6pt}
\begin{tabular}{l|c|c|c}
\hline
 & Small   & Medium    &Large  \\ 
\hline
    TDRG~\cite{tdrg}   &31.3   &69.4   &85.1   \\
    Q2L~\cite{liu2021query2label}   &30.7&   70.2&   85.6\\ 
    GKGNet  & \textbf{35.6} & \textbf{73.6}  & \textbf{86.6}    \\
     \hline
\end{tabular}%
}
\label{tab:different-scale-map}
\end{minipage}
\hspace{1.5mm}
\begin{minipage}[t]{0.5\textwidth}
\centering
\caption{\textbf{Sensitivity to random initial values}. We report results on MS-COCO with $576 \times 576$ input size.
}
\scriptsize 
\centering
\resizebox{ 0.98\linewidth}{!}{%
\renewcommand{\arraystretch}{1.8}
\begin{tabular}{l|c|c|c|c|c}
\hline
\multirow{2}{*}{} & \multicolumn{3}{c|}{All} & \multicolumn{2}{c}{Top3} 
 \\ \cline{2-6}
 & mAP   & CF1    & OF1  & CF1 & OF1  \\ 
    \hline
    {GKGNet}    & $87.7\pm 0.0$   & $82.2 \pm 0.1$   & $83.8 \pm 0.1$  &$77.6 \pm 0.0$     & $79.3 \pm 0.0$\\
    \hline
\end{tabular}%
}
\label{tab:sensitivity}
\end{minipage}
\end{table*}

\subsubsection{Effect of object scales.}
MS-COCO   divides objects into small (area $< 32^2$), medium ($32^2 <$ area $< 96^2$) and large (area $> 96^2$). Results for various target object scales are shown in Table~\ref{tab:different-scale-map}, with GKGNet achieving a notable 4.9\% mAP gain for small objects compared to Q2L.
Q2L faces challenges with small objects due to excessive whole-image interactions and background interference in transformers. In contrast, GKGNet selects regions of interest, saving computation and bypassing background disruptions.

\subsubsection{Sensitivity to random initial values.} 
We conduct experiments with three different random seeds for initialization and report their performances via mean and variance. Experiments show that the model's performance is robust to different initial values.

\begin{figure}[tb]
  \centering
  \begin{subfigure}[b]{0.43\textwidth}
    \centering
    \begin{tikzpicture}[scale=0.85]
        \begin{axis}[            width=1.2\textwidth,            height=0.55\textwidth,            xlabel={$G$},            ylabel={mAP},            xmin=0, xmax=9,            ymin=84, ymax=87,            xtick={1,2,4,8},            ytick={85, 86, 87},            legend style={at={(0.5,-0.15)},anchor=north}, 
        ymajorgrids=true,          
        grid style=dashed,   
        axis x line=bottom,
axis y line=left,
axis line style = {-latex},
every axis x label/.style={at={(ticklabel* cs:1.15)},anchor=east},
every axis y label/.style={at={(ticklabel* cs:1.5)},anchor=north },
        ]
        \addplot[            color=black,            mark=*,            ]
            coordinates {
            (1,85.5)(2,86.7)(4,86.4)(8,86.5)
            };
        \end{axis}
    \end{tikzpicture}
    \label{fig:table1}
  \end{subfigure}
  \begin{subfigure}[b]{0.43\textwidth}
    \centering
    \begin{tikzpicture}[scale=0.85]
        \begin{axis}[            width=1.2\textwidth,            height=0.55\textwidth,            xlabel={$K$},            ylabel={mAP},            xmin=2, xmax=20,            ymin=84, ymax=87,            xtick={3,6,9,12,15,18},            ytick={ 85, 86, 87},            legend style={at={(0.5,-0.15)},anchor=north},            ymajorgrids=true,            grid style=dashed, 
                axis x line=bottom,
        axis y line=left,
axis line style = {-latex},
every axis x label/.style={at={(ticklabel* cs:1.15)},anchor=east},
every axis y label/.style={at={(ticklabel* cs:1.5)},anchor=north },]
        \addplot[            color=black,            mark=*,            ]
            coordinates {
            (3,84.5)(6,86.5)(9,86.7)(12,86.6)(15,86.5)(18,86.4)
            };
        \end{axis}
    \end{tikzpicture}
    \label{fig:table2}
  \end{subfigure}
  \caption{Effect of the number of groups $G$ (\textbf{Left}) and number of neighbors $K$ (\textbf{Right}).}
  \label{fig:effect-of-G-k}
\end{figure}
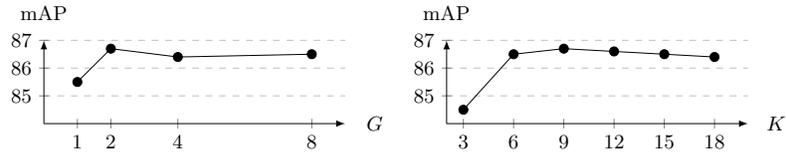

\subsubsection{Effect of $G$ and $K$.} 
For Group KGCN module, the number of groups $G$ and the neighbor number $K$ are important hyper-parameters. We explore the effect of $G$ from 1 to 8 and show the results in Fig.~\ref{fig:effect-of-G-k} (Left). Setting $G = 2$ significantly improves the performance upon $G = 1$ (traditional KNN), which validates the importance of dynamic neighbor number choosing. Further increasing $G$ does not bring more gains, so we choose $G = 2$ for simplicity.
For neighbor number $K$, as shown in Fig.~\ref{fig:effect-of-G-k} (Right),
having too few neighbors (\eg $K=3$) leads to the loss of information, while larger $K$ performs similarly well.
We set $K=9$, aligning with ViG.

\begin{figure}[t]\centering
\includegraphics[width=0.7\textwidth]{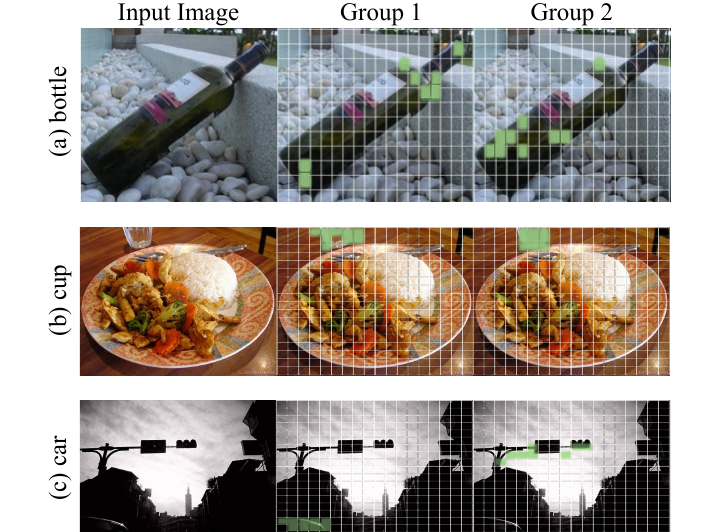}
\caption{Visualization of the learned connections between label node and patch nodes in the Cross-Level Group KGCN module. The colored blocks indicate that the patches are connected to the label ``bottle'', ``cup'', or ``car''.
}
\label{fig:visualization-label-patch}
\end{figure}

\begin{figure}[tb]\centering
\includegraphics[width=0.7\textwidth]{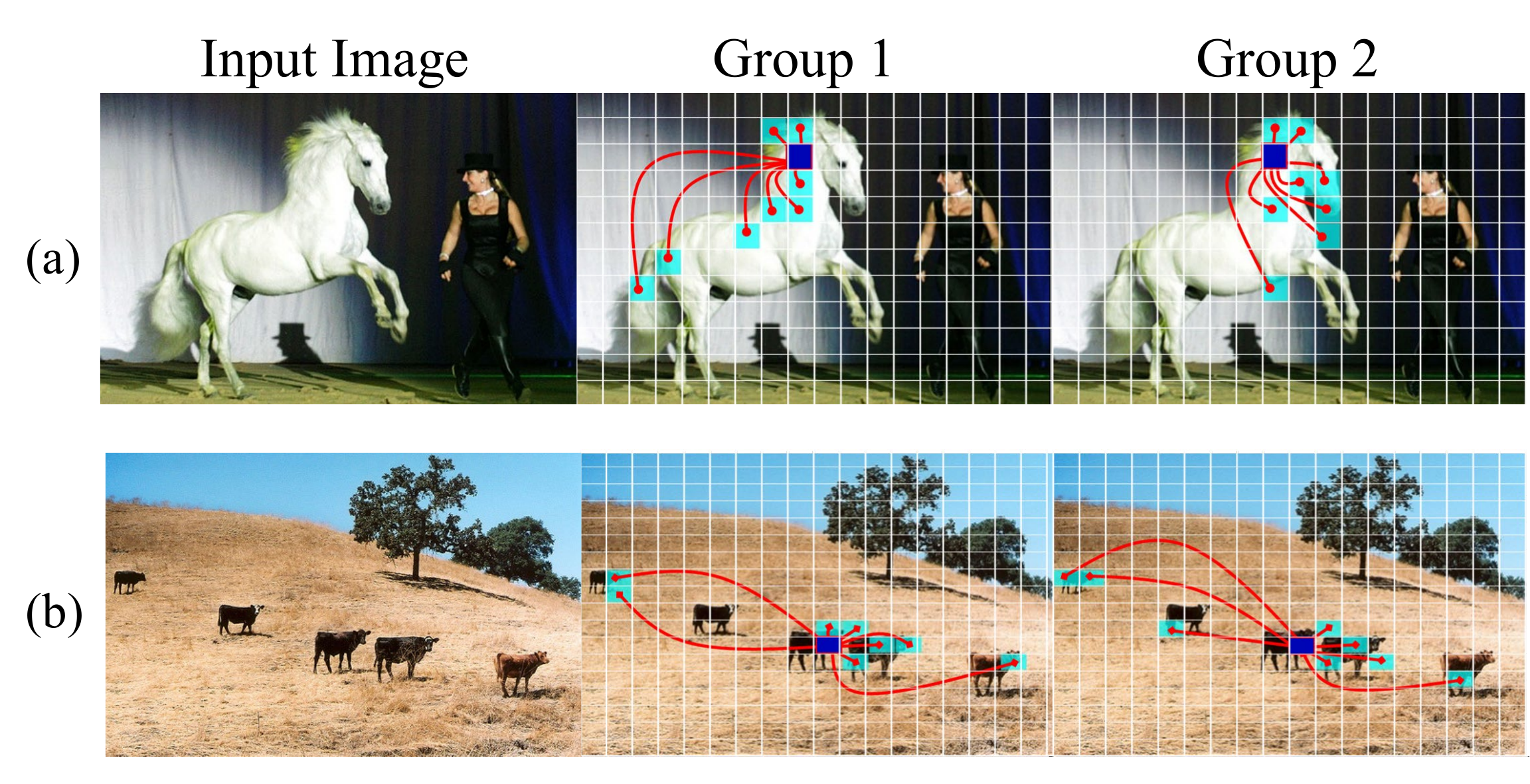}
\caption{Visualization of connections in the Patch-Level Group KGCN module. Deep blue represents the destination node, and baby blue patches are its selected neighbors. Red lines depict the connections between patch nodes.
}
\label{fig:visualization-patch-patch}
\end{figure}

\subsection{Visualization and Analysis}
\label{sec:visualization}

\textbf{Cross-Level Group KGCN module.}
In Fig.~\ref{fig:visualization-label-patch}, we visualize the connections of different groups at the last stage of Cross-Level Group KGCN module. Cross-Level Group KGCN module adaptively connects each label embedding to patches of different scales, shapes and locations.
In Fig.\ref{fig:visualization-label-patch} (a), for the label ``bottle'',
one group focuses on the bottom and neck, and the other captures the body, sufficiently avoiding information loss.
In Fig.\ref{fig:visualization-label-patch} (b), for the smaller ``cup'', different groups select several overlapped patches, effectively capturing foreground regions while reducing background interference.
In Fig.~\ref{fig:visualization-label-patch} (c), the Group KGCN module captures co-occurring categories through multiple perspectives. For the lable ``car'', the first group identifies car areas, while the second group locates the area of traffic lights, enhancing semantic relations and producing robust predictions.

\textbf{Patch-Level Group KGCN module.}
Patch-Level Group KGCN module updates features of each patch by interacting with neighbor patches to capture semantic visual relationships flexibly.
In Fig.\ref{fig:visualization-patch-patch} (a), the destination patch (deep blue) selects two different groups of  source patches (baby blue) to interact with different parts of the horse: forequarters and hindquarters. It learns message passing among semantically related patches, covering regions of interest through multiple groups. In Fig.\ref{fig:visualization-patch-patch} (b), the destination patch selects spatially distributed foreground patches (multiple cows) as neighbors, enabling long-range correlation. The overlapped selection of two groups avoids background interference due to the small size of foreground areas.

\section{Conclusion}
In this paper, we propose GKGNet, a novel fully graph convolutional model for the task of MLIR. We pioneer to study the unified graph representations for both visual features and label embeddings. Group KGCN module is proposed for dynamic graph construction and message passing. It is effective in handling the scale variance of different objects, capturing information from different perspectives, and modeling the co-occurrence of different objects.  
Comprehensive experiments on public benchmark datasets, \ie MS-COCO and VOC2007, demonstrate the effectiveness of our method. 
We hope the idea of unifying multi-modality features with dynamic graph representations can be broadly useful and our work can draw the community's attention to this promising direction.
In the future, we plan to extend our work to a wider range of graph-based learning problems, \eg point clouds and social networks.

\textbf{Limitations.}
Due to resource limitation, we are not able to present results of larger-scale models and extensive pre-trainin, \eg ImageNet22K, which requires thousands of GPU hours. We're still actively seeking collaborations and more resources to scale up both the model and data.

\noindent\textbf{Acknowledgement.}
This paper is partially supported by the National Key R\&D Program of China No.2022ZD0161000 and the General Research Fund of Hong Kong No.17200622 and 17209324.

%
%
\bibliographystyle{splncs04}
\bibliography{egbib}

\clearpage
\appendix

\setcounter{table}{0}
\renewcommand{\thetable}{A\arabic{table}}
\setcounter{figure}{0}
\renewcommand{\thefigure}{A\arabic{figure}}

\section{Implementation Details in General Image Classification}

In Table 5 of the main text, we demonstrate the efficiency of Group KNN in general image classification, highlighting its robust performance in graph construction. We maintain uniform training settings for both models, with the only difference being the use of our proposed Group KNN.

We adhere to widely-used settings for each dataset, and specific details for each dataset are provided below:

\paragraph{ImageNet-1K}
We follow the training settings of ViG~\cite{vig}, keeping hyperparameters consistent. Data augmentation involves Mixup and CutMix techniques, and input images are resized to $224\times 224$. The AdamW optimizer is used with parameters: learning rate ($lr$) of 0.002, weight decay of 0.05, and a cosine learning rate update policy. The training process spans up to a maximum of 300 epochs.

\paragraph{CIFAR-10 and CIFAR-100}
Given our adoption of $K=9$ for KNN graph construction, the standard $32\times 32$ input size in CIFAR-10 and CIFAR-100 doesn't provide sufficient patches in the last three stages. Therefore, we modify the input size to $224\times 224$, consistent with ImageNet-1K. The remaining training configurations align with MMClassification~\cite{mmclassification}. Data augmentation includes image resizing and random flipping. We use the SGD optimizer with parameters: learning rate ($lr$) of 0.1, weight decay of 0.0001, and a step learning rate update policy that reduces the learning rate at epochs 100 and 150. The training process continues for a maximum of 200 epochs.

\paragraph{Flowers}
We set the image size to $224\times 224$. Data augmentation includes image resizing and random flipping. The SGD optimizer is used with parameters: learning rate ($lr$) of 0.1, weight decay of 0.0001, and a step learning rate update policy that reduces the learning rate at epochs 30, 60, and 90. The training process spans up to a maximum of 100 epochs.

\begin{figure}[htb]
\begin{center}
\includegraphics[width=0.87\linewidth]{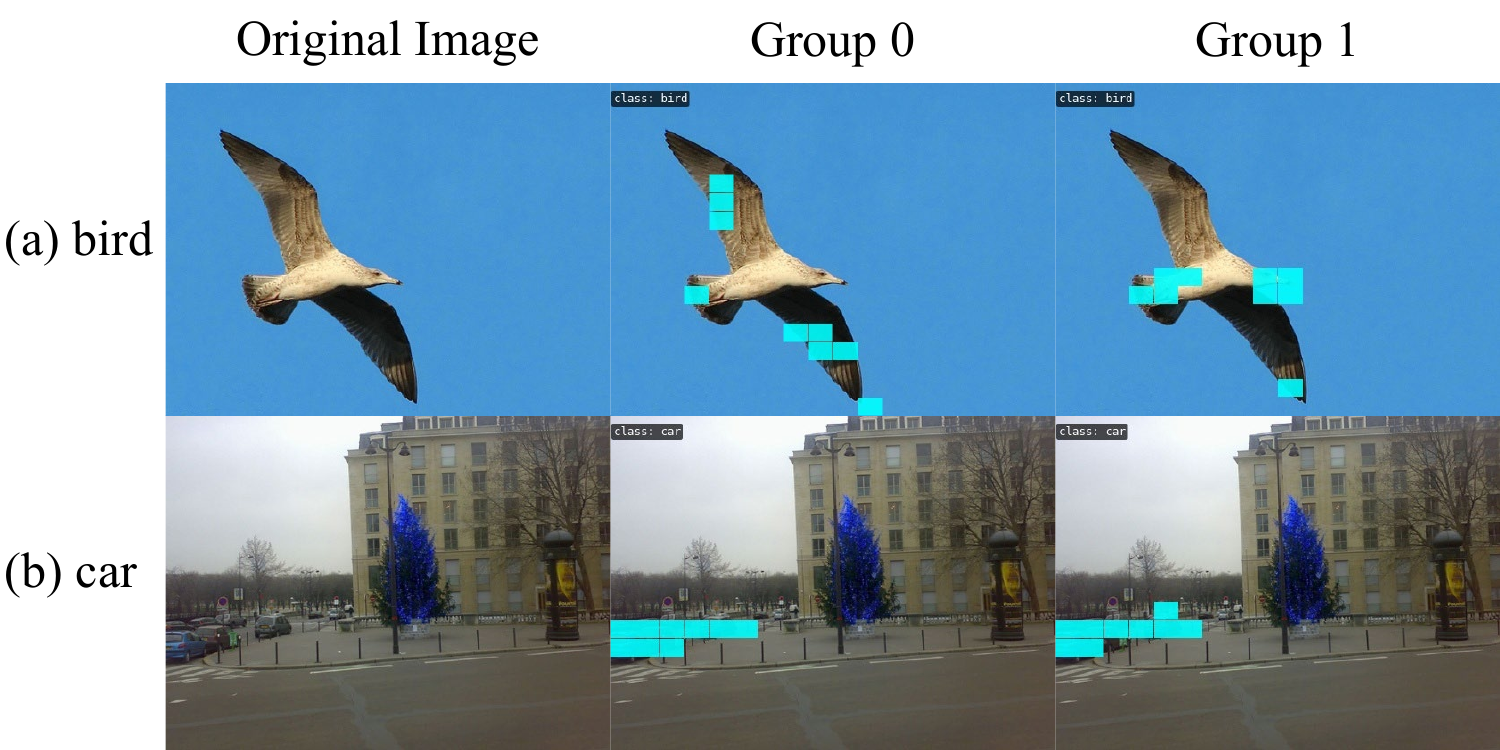}
\end{center}
\caption{Visualization of the learned connections between patch nodes and label nodes for \textbf{objects of varying sizes}. (a) and (b) correspond to ``bird" and ``car" respectively. The azure-colored blocks in the figure are the patch nodes nearest to the label nodes.}
\label{fig:vis-varying-size-object}
\end{figure}

\begin{figure}[htb]
\begin{center}
\includegraphics[width=0.99\linewidth]{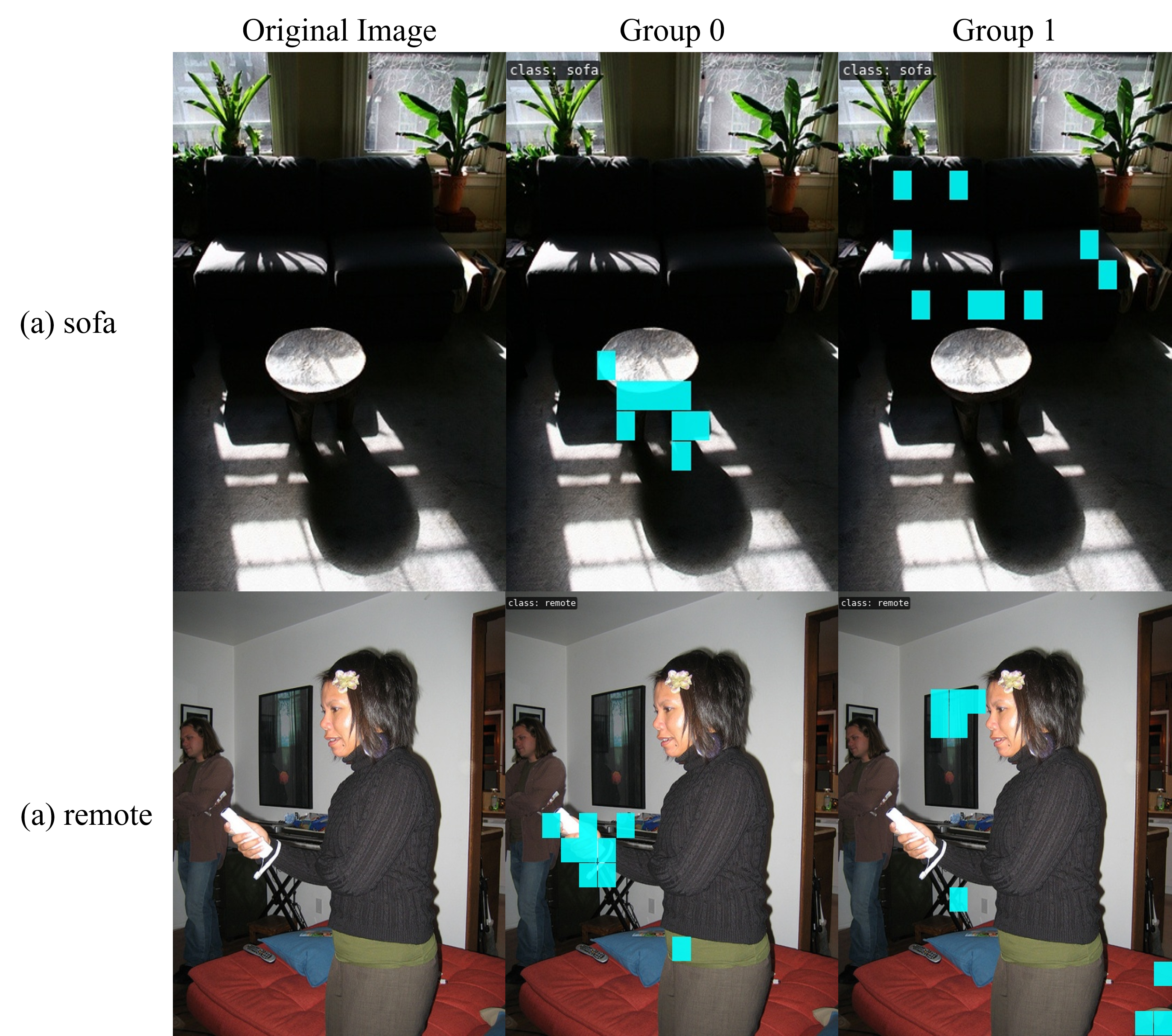}
\end{center}
\caption{Visualization of the learned connections between patch nodes and their label nodes for \textbf{co-occurring categories}. (a) and (b) correspond to ``sofa'' and ``remote'' respectively. The two groups of nodes have found the patches corresponding to their own and their correlated classes respectively.}
\label{fig:vis-co-occurring-categories}
\end{figure}

\begin{figure}[htb]
\begin{center}
\includegraphics[width=0.99\linewidth]{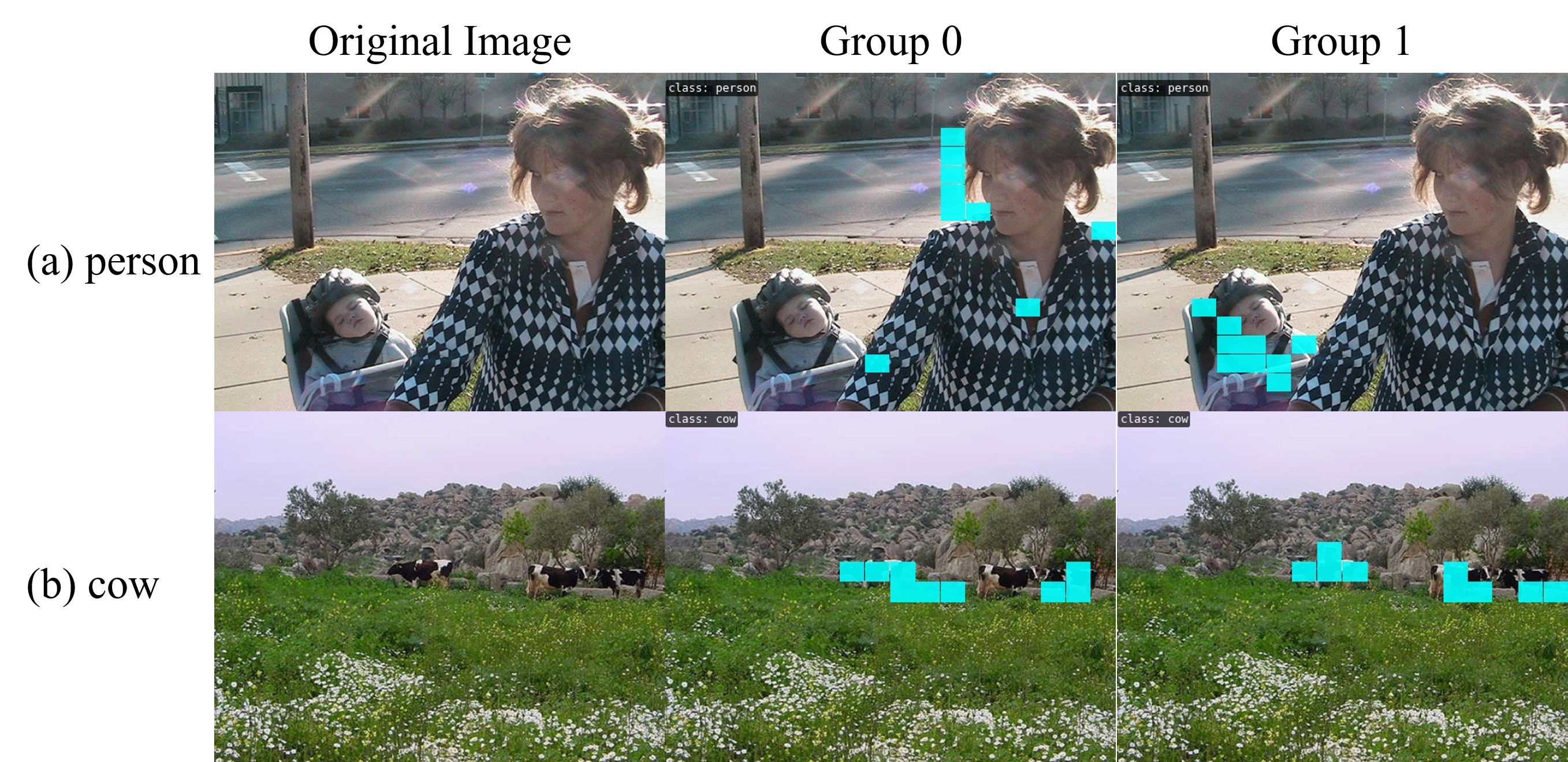}
\end{center}
\caption{Visualization of the learned connections between patch nodes and label nodes for \textbf{spatially distributed objects}. (a) and (b) correspond to ``person'' and ``cow'' respectively. The two groups of nodes completely found multiple objects related to the label in both (a) and (b).}
\label{fig:vis-multiple-objects-of-one-label}
\end{figure}

\begin{figure}[htb]
\begin{center}
\includegraphics[width=0.9\linewidth]{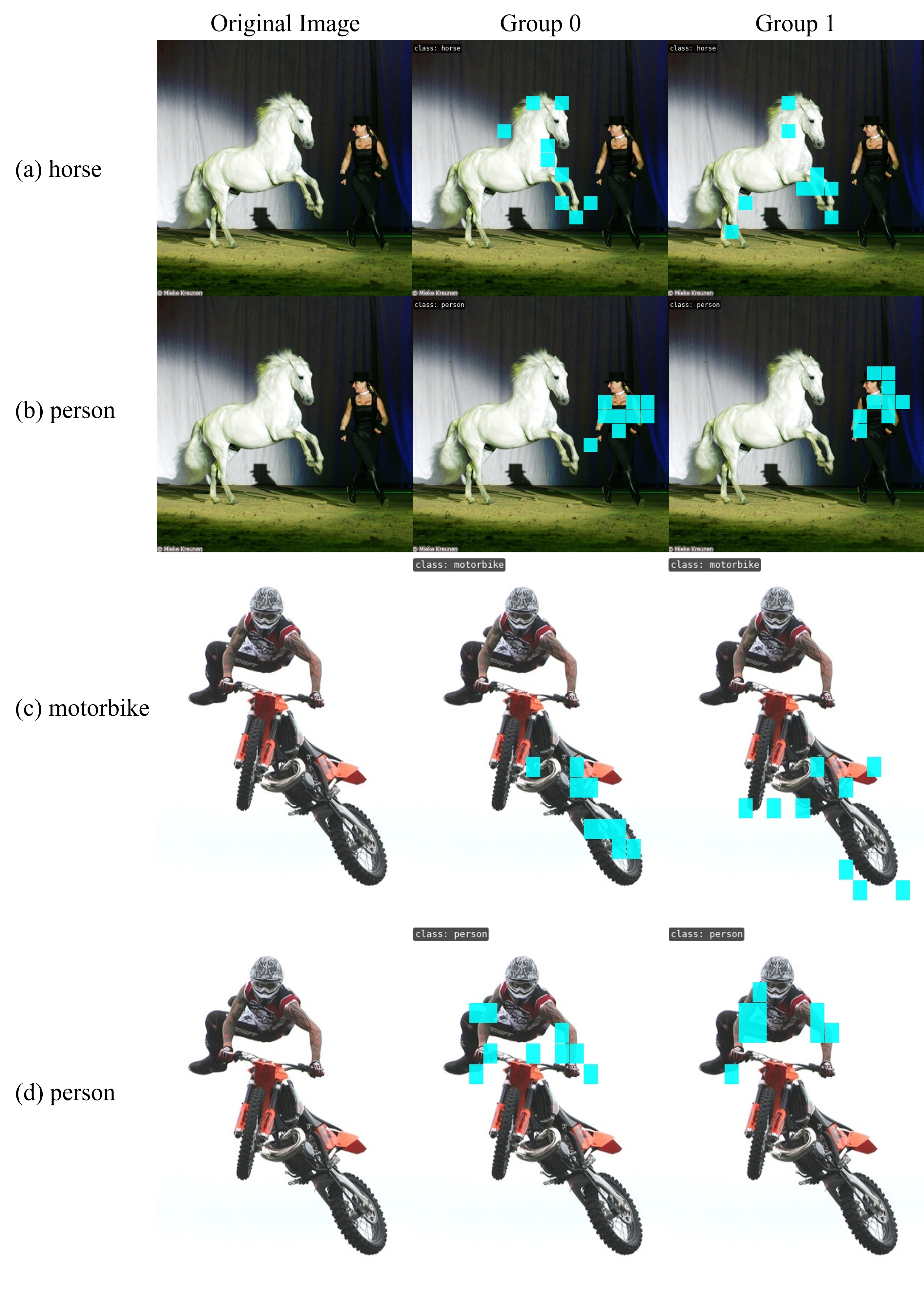}
\end{center}
\caption{Visualization of the learned connections between patch nodes and label nodes, where there are \textbf{multiple labels in one image}. Every two figures refer to two labels in the same image. Each label node has found its correct corresponding patch areas in the case where multiple classes exist simultaneously.}
\label{fig:vis-multiple-labels-of-one-image}
\end{figure}

\section{More Qualitative Analysis}
\subsection{Learned connections between the label node and the patch nodes in Group KGCN module}
\label{sec:more-vis-patch-label}
In the section, we will show more visualization of the learned connections between patch nodes and label nodes in the last GKG block to demonstrate the effectiveness of our model on the MLIR task.

\paragraph{Objects of varying sizes.}
As shown in Fig.~\ref{fig:vis-varying-size-object} (a),  ``bird''   occupies a large region in images, and using a fixed small number of $K$ can hardly cover enough patches to obtain sufficient distinguished features. However, our Group KGCN module learns to pay attention to different areas in different groups, \eg  the body and wings . The label   ``bird''   selects $17$ different patch nodes, which are close to the maximum $K \times G = 18$ patches, to utilize features from enough parts and avoid information loss.
As shown in Fig.~\ref{fig:vis-varying-size-object} (b),   ``car'' only occupies a small area in the images. If we extract global features by global pooling in CNNs or global attention in Transformers, most features will come from the background regions which will hinder feature representation learning. In contrast, our Group KGCN learns to select two groups of neighbors that have a large overlap with each other, focusing on the foreground objects and eliminating background distractions. As a result, the label ``car'' selects $10$ different patch nodes, which are close to the minimum $K=9$ patches, to avoid background interference.

\paragraph{Co-occurring Categories.} Modeling multi-label correlations are critical for MLIR tasks, especially when the corresponding areas of the target label are occluded or unclear. As shown in Fig.~\ref{fig:vis-co-occurring-categories}(a), the neighbor patch nodes of ``sofa'' in the first group precisely locate the sofa which is in the dark, and the second group pays attention to the table in front of the sofa. 
In Fig.~\ref{fig:vis-co-occurring-categories}(b), the first group of ``remote'' concentrates on the area partly covered by a person's hand where the remote is located, while the other group focuses on the TV. 
Capturing the features from co-occurring labels with our multi-group design will provide more robust relevant information which helps to understand the whole scene.

\paragraph{Spatially Distributed Objects.} 
Objects are commonly spatially distributed in the image. As shown in Fig.~\ref{fig:vis-multiple-objects-of-one-label}, there are two people in Fig.~\ref{fig:vis-multiple-objects-of-one-label}(a) and three cows in Fig.~\ref{fig:vis-multiple-objects-of-one-label}(b), and all objects corresponding to the target labels are selected by two groups, which demonstrates our Group KGCN module can effectively extract features from the long-range area. 

\paragraph{Multiple labels.}
For MLIR, it is very common that multiple labels occur in a single image. As shown in Fig.~\ref{fig:vis-multiple-labels-of-one-image}, both two images have two categories. And each label node exactly finds its related image patches without confusion, which validates the effect of our fully graph network in unifying the representations of patch nodes and label nodes.

\section{Detailed Results on MS-COCO Dataset}
In Table~\ref{tab:ms-coco-detale-ap}, we report and compare the per-category recognition accuracy on MS-COCO dataset~\cite{ms-coco}. We observe that our model achieves the best performance for 76 out of 80 categories and achieve very competitive results for the rest 4 categories. Especially, for small and challenging object categories, \eg hair drier, scissors, apple and toothbrush, the performance improvements are more significant. 

\begin{table*}[htpb]
  \centering
  \resizebox{\linewidth}{!}{
  \centering
    \begin{tabular}{c|c|c|c|c|c|c|c|c}
    \hline\hline
          & person & bicycle & car   & motorcycle & airplane & bus   & train & truck \\
           \hline
    TDRG & 99.2  & 81.6  & 89.9  & 94.6  & 97.1  & 89.6  & 96.7  & 77.1 \\
    Q2L & \textbf{99.3} & 81.9  & 89.9  & 94.4  & 97.5  & 89.7  & 96.7  & 77.3 \\
    GKGNet & \textbf{99.3} & \textbf{88.9} & \textbf{93.2} & \textbf{96.4} & \textbf{98.0} & \textbf{91.8} & \textbf{98.1} & \textbf{81.4} \\
    \hline\hline
          & boat  & traffic light & fire hydrant & stop sign & parking meter & bench & bird  & cat \\
           \hline
    TDRG & 91.3  & 86.6  & 85.8  & 80.3  & 71.8  & 69.0 & 83.6  & 96.8 \\
    Q2L & 91.1  & 86.9  & 86.0 & 81.1  & 72.7  & 70.2  & 84.0 & 97.1 \\
    GKGNet & \textbf{93.2} & \textbf{91.6} & \textbf{90.4} & \textbf{85.8} & \textbf{77.2} & \textbf{74.5} & \textbf{87.5} & \textbf{97.2} \\
    \hline\hline
          & dog   & horse & sheep & cow   & elephant & bear  & zebra & giraffe \\
           \hline
    TDRG & 91.2  & 94.4  & 96.5  & 93.3  & 98.6  & 98.0 & 99.4  & 99.7 \\
    Q2L & 92.0 & 95.3  & 96.5  & 93.9  & 98.8  & \textbf{98.6} & 99.5  & \textbf{99.8} \\
    GKGNet & \textbf{94.6} & \textbf{96.3} & \textbf{96.8} & \textbf{94.8} & \textbf{99.0} & 97.8  & \textbf{99.6} & \textbf{99.8} \\
    \hline\hline
          & backpack & umbrella & handbag & tie   & suitcase & frisbee & skis  & snowboard \\
           \hline
    TDRG & 55.2  & 87.4  & 57.8  & 86.1  & 76.5  & 94.9  & 95.3  & 87.4 \\
    Q2L & 54.1  & 88.2  & 57.7  & 86.2  & 77.6  & 94.9  & 96.1  & 87.0 \\
    GKGNet & \textbf{59.4} & \textbf{90.5} & \textbf{62.9} & \textbf{88.1} & \textbf{81.9} & \textbf{97.1} & \textbf{96.6} & \textbf{89.5} \\
    \hline\hline
          & sports ball & kite  & baseball bat & baseball glove & skateboard & surfboard & tennis racket & bottle \\ \hline
    TDRG & 88.8  & 97.6  & 95.4  & 96.1  & 97.6  & 96.7  & 98.9  & 76.3 \\
    Q2L & 89.2  & 98.3  & 96.5  & \textbf{96.7} & 98.0 & 96.5  & 99.2  & 76.9 \\
    GKGNet & \textbf{92.4} & \textbf{98.6} & \textbf{96.6} & 96.5  & \textbf{98.5} & \textbf{97.8} & \textbf{99.3} & \textbf{82.1} \\
    \hline\hline
          & wine glass & cup   & fork  & knife & spoon & bowl  & banana & apple \\ \hline
    TDRG & 80.9  & 78.1  & 80.3  & 71.4  & 67.5  & 76.2  & 87.5  & 71.4 \\
    Q2L & 80.6  & 79.3  & 82.4  & 72.9  & 68.5  & 76.4  & 87.7  & 72.3 \\
    GKGNet & \textbf{86.7} & \textbf{82.9} & \textbf{85.7} & \textbf{77.4} & \textbf{72.1} & \textbf{81.4} & \textbf{92.9} & \textbf{79.9} \\
    \hline\hline
          & sandwich & orange & broccoli & carrot & hot dog & pizza & donut & cake \\ \hline
    TDRG & 77.3  & 82.5  & 93.2  & 81.5  & 80.5  & 95.1  & 86.7  & 84.9 \\
    Q2L & 78.2  & 83.3  & 93.8  & 82.2  & 82.0 & 95.6  & 86.0 & 84.6 \\
    GKGNet & \textbf{81.8} & \textbf{88.3} & \textbf{94.5} & \textbf{86.3} & \textbf{85.1} & \textbf{95.9} & \textbf{88.9} & \textbf{88.2} \\
    \hline\hline
          & chair & couch & potted plant & bed   & dining table & toilet & tv    & laptop \\ \hline
    TDRG & 80.8  & 84.9  & 71.6  & 89.0 & 80.4  & 97.8  & 89.5  & 91.0 \\
    Q2L & 82.0 & 85.5  & 72.7  & 89.9  & 82.1  & 98.0 & 90.5  & 90.6 \\
    GKGNet & \textbf{84.6} & \textbf{87.0} & \textbf{76.9} & \textbf{90.5} & \textbf{82.2} & \textbf{98.2} & \textbf{91.5} & \textbf{93.1} \\
    \hline\hline
          & mouse & remote & keyboard & cell phone & microwave & oven  & toaster & sink \\ \hline
    TDRG & 89.9  & 82.2  & 89.2  & 70.0 & 82.4  & 88.8  & \textbf{40.8} & 92.2 \\
    Q2L & 90.6  & 82.2  & 89.9  & 70.0 & \textbf{84.9} & 90.1  & 35.7  & 92.6 \\
    GKGNet & \textbf{92.2} & \textbf{86.4} & \textbf{91.3} & \textbf{74.5} & 84.7  & \textbf{90.7} & 40.0 & \textbf{93.4} \\
    \hline\hline
          & refrigerator & book  & clock & vase  & scissors & teddy bear & hair drier & toothbrush \\ \hline
    TDRG & 82.8  & 70.7  & 84.2  & 79.9  & 65.5  & 88.0 & 45.1  & 76.5 \\
    Q2L & 82.7  & 71.7  & 84.6  & 81.0 & 68.1  & 89.2  & 43.5  & 75.4 \\
    GKGNet & \textbf{86.4} & \textbf{77.1} & \textbf{87.6} & \textbf{83.0} & \textbf{76.4} & \textbf{91.1} & \textbf{62.0} & \textbf{80.6} \\
    \hline\hline
    \end{tabular}%
    }
  \caption{AP for each category obtained by competing methods on MS-COCO dataset. The best scores are highlighted in bold.}
  \label{tab:ms-coco-detale-ap}%
\end{table*}%

\end{document}

%% file: tables/table_coco.tex
\begin{table*}[t]
\caption{\textbf{Comparisons} with state-of-the-art methods on MS-COCO. All the methods adopt models pre-trained on ImageNet-1K dataset. $^\dagger$ means using model EMA.
We report multiple evaluation metrics (higher is better), among which mAP, CF1, and OF1 are the primary ones.
GKGNet significantly outperforms the existing approaches in terms of both accuracy and efficiency. 
}
\scriptsize 
\centering
\resizebox{0.98\linewidth}{!}{%
\begin{tabular}{l|c|c|c|a|c|c|a|c|c|a|a|a}
\hline
    \multirow{2}{*}{} &
    \multirow{2}{*}{} &
    \multirow{2}{*}{} &
    \multirow{2}{*}{} &
  \multicolumn{6}{c}{All} &
  \multicolumn{1}{c|}{} &
  \multicolumn{2}{c}{Top3} 
 \\ \cline{5-13}
&Resolution	& Param(M)	& FLOPs(G) & mAP  & CP   & CR   & CF1  & OP   & OR   & OF1  & CF1 & OF1  \\ 
\hline\hline
    {ResNet-101~\cite{resnet101}} & $224\times 224$   & 44.5  & 7.8  & 78.3  & 80.2  & 66.7  & 72.8  & \textbf{83.9}  & 70.8  & 76.8  & 69.7  & 73.6 \\
    {SRN~\cite{srn}} & $224\times 224$   & 76.8     & 9.0     & 77.1  & 81.6  & 65.4  & 71.2  & 82.7  & 69.9  & 75.8  & 67.4  & 72.9 \\
    {{IDA-SwinS(H)~\cite{liu2022causality}}} & $224\times 224$   & 48.9 & 8.7     & 80.6  &64.5  &\textbf{81.1}   &71.8    &65.8    &\textbf{83.9}   &73.8   &71.4      &74.5  \\
    {Mltr~\cite{mltr}} & $224\times 224$   & \textbf{33.0}  & -     & 81.9  & 80.7  & 71.5  & 75.2  & 81.4  & {76.3}  & 78.1  & -     & - \\
    {GKGNet (Ours)} & $224\times 224$  &     33.3  & \textbf{5.2}   & \textbf{82.0} & \textbf{81.7} & {73.1} & \textbf{77.1} & 83.1  & {76.3} & \textbf{79.6} & \textbf{73.7} & \textbf{76.1} \\
     \hline\hline
     {CADM~\cite{cadm}} & $448\times 448$   & -     & -     & 82.3  & 82.5  & 72.2  & 77.0  & 84.0  & 75.6  & 79.6  & 73.5  & 76.0 \\
     {ML-GCN~\cite{ml-gcn}} & $448\times 448$   & 44.9  & 31.5      & 83.0  & 85.1  & 72.0  & 78.0  & 85.8  & 75.4  & 80.3  & 74.6  & 76.7 \\
     {KSSNet~\cite{kssnet}} & $448\times 448$   & 173.8 & -     & 83.7  & 84.6  & 73.2  & 77.2  & 87.8  & 76.2  & 81.5  & -     & - \\
     {MS-CMA~\cite{ms-cma}} & $448\times 448$   & -     & -     & 83.8  & 82.9  & 74.4  & 78.4  & 84.4  & 77.9  & 81.0  & 74.9  & 77.1 \\
     {MCAR~\cite{mcar}} & $448\times 448$   & -     & -     & 83.8  & 85.0  & 72.1  & 78.0  & \textbf{88.0}  & 73.9  & 80.3  & 75.1  & 76.7 \\
     {TDRG~\cite{tdrg}} & $448\times 448$   & 68.3  & 42.2     & 84.6  & {86.0}  & 73.1  & 79.0  & 86.6  & 76.4  & 81.2  & 75.0  & 77.2 \\
     {Q2L-R101~\cite{liu2021query2label}$^\dagger$} & $448\times 448$   & 193.6 & 51.4      & 84.9  & 84.8  & 74.5  & 79.3  & 86.6  & 76.9  & 81.5  & 73.3  & 75.4 \\
     {{IDA-SwinS(H)~\cite{liu2022causality}}} & $448\times 448$   & 48.9 & 35.0     & 85.5  &68.9  &\textbf{85.6}   &76.3    &69.8    &\textbf{87.5}   &77.7   &75.2      &77.4  \\
     {GKGNet (Ours)} & $448\times 448$   &\textbf{34.0} & \textbf{21.9} & \textbf{86.7} & \textbf{86.4} & {77.1} & \textbf{81.5} & 87.3  & {79.7} & \textbf{83.3} & \textbf{77.0} & \textbf{78.8} \\
     \hline\hline
     {SSGRL~\cite{ssgrl}} & $576\times 576$   & 92.3  & 68.5      &  83.8  & \textbf{91.9}  & 62.5  & 72.7  & \textbf{93.8}  & 64.1  & 76.2  & 76.8  & \textbf{79.7} \\
     {KGGR~\cite{chen2020knowledge}} & $576\times 576$   & -  & -   &  84.3  & 85.6  & 72.7  & 78.6  & 87.1  & 75.6  & 80.9  & 75.0  & 77.0 \\
     {MCAR~\cite{mcar}} & $576\times 576$   & -     & -     & 84.5  & 84.3  & 73.9  & 78.7  & 86.9  & 76.1  & 81.1  & 75.3   & 77.0 \\
     {ADD-GCN~\cite{add-gcn}} & $576\times 576$   & 48.2  & 52.7    & 85.2  & 84.7  & 75.9  & 80.1  & 84.9  & 79.4  & 82.0  & 75.8  & 77.9 \\
     {C-Tran~\cite{c-tran}} & $576\times 576$   & 120.4 & 84.2  & 85.1  & 86.3  & 74.3  & 79.9  & 87.7  & 76.5  & 81.7  & 76.0  & 77.6 \\
     {TDRG~\cite{tdrg}} & $576\times 576$   & 68.3  & 69.8      & 86.0  & 87.0  & 74.7  & 80.4  & 87.5  & 77.9  & 82.4  & 76.2  & 78.1 \\
    {{IDA-R101(H)~\cite{liu2022causality}}} & $576\times 576$  & 53.2 & 54.1 & 86.3  & -  & -   &80.4    &- &-   &82.5   &76.4      &78.2  \\
    {{IDA-SwinS(H)~\cite{liu2022causality}}} & $576\times 576$  & 48.9 & 64.0     & 86.4  &69.6  &\textbf{86.5}   &77.1    &69.6    &\textbf{88.8}   &78.1   &75.6      &77.8 \\
     {Q2L-R101~\cite{liu2021query2label}$^\dagger$} & $576\times 576$   & 193.6 & 80.8  & 86.5  & 85.8  & 76.7  & 81.0  & 87.0  & 78.9  & 82.8  & 76.5  & 78.3 \\
     {GKGNet (Ours)} & $576\times 576$   & \textbf{34.7}  & \textbf{40.1} & \textbf{87.7} & 87.0  & {78.5} & \textbf{82.5} & 87.6  & {81.0} & \textbf{84.2} & \textbf{77.6} & 79.3 \\
    \hline
\end{tabular}%
}
\label{tab:ms-coco-table}
\end{table*}

%% file: tables/table_else.tex
\begin{table*}[t]
\caption{\textbf{Comparisons} with state-of-the-art methods on Pascal VOC2007 dataset. We report the average precision in each category, and the mean average precision (mAP) of all the categories. All the models are pre-trained on the MS-COCO  ($576 \times 576$ input size). Our proposed GKGNet outperforms the previous state-of-the-arts.
}
\label{tab:voc2007-table}
\begin{center}
\renewcommand{\arraystretch}{1.2}
\resizebox{\linewidth}{!}{
\begin{tabular}{l||c|c|c|c|c|c|c|c|c|c|c|c|c|c|c|c|c|c|c|c||a}
\hline
  \cline{2-22}\cline{2-22}
&{Aero} & {Bike} & {Bird} & {Boat} & {Bottle} & {Bus} & {Car} & {Cat} & {Chair} & {Cow} & {Table} & {Dog} & {Horse} & {Mbike} & {Person} & {Plant} & {Sheep} & {Sofa} & {Train} & {TV} & {mAP} \\  
\hline\hline
    {SSGRL} & 99.7  & 98.4  & 98.0  & 97.6  & 85.7  & 96.2  & 98.2  & 98.8  & 82.0  & 98.1  & 89.7  & 98.8  & 98.7  & 97.0  & 99.0  & 86.9  & 98.1  & 85.8  & 99.0  & 93.7  & 95.0 \\
    {ASL} & \textbf{99.9}  & 98.4  & 98.9  & 98.7  & 86.8  & 98.2  & 98.7  & 98.5  & 83.1  & 98.3  & 89.5  & 98.8  & 99.2  & 98.6  & 99.3  & 89.5  & \textbf{99.4}  & 86.8  & {99.6}  & 95.2  & 95.8 \\
    {ADD-GCN} & 99.8  & 99.0  & 98.4  & 99.0  & 86.7  & 98.1  & 98.5  & 98.3  & 85.8  & 98.3  & 88.9  & 98.8  & 99.0  & 97.4  & 99.2  & 88.3  & 98.7  & \textbf{90.7}  & 99.5  & 97.0  & 96.0 \\
    {Q2L} & \textbf{99.9}  & 98.9  & 99.0  & 98.4  & \textbf{87.7}  & \textbf{98.6}  & {98.8}  & 99.1  & 84.5  & 98.3  & 89.2  & 99.2  & 99.2  & \textbf{99.2}  & 99.3  & \textbf{90.2}  & {98.8}  & 88.3  & 99.5  & 95.5  & 96.1 \\
    {GKGNet (Ours)}  &\textbf{99.9} & \textbf{99.4} & \textbf{99.2} & \textbf{99.4} & 87.0  & 98.2  & \textbf{99.1} & \textbf{99.6} & \textbf{88.4} & \textbf{99.5} & \textbf{92.6} & \textbf{99.5} & \textbf{99.5} & 98.7  & \textbf{99.5} & 89.5  & 99.3  & 90.4  & \textbf{99.7} & \textbf{97.2} & \textbf{96.8} 
    \\
    \hline
\end{tabular}}
\end{center}
\end{table*}